\documentclass[3p,times,twocolumn,11pt, procedia]{elsarticle}
\usepackage{ecrc}
\usepackage[utf8]{vietnam}
\usepackage[hang,flushmargin]{footmisc}
\usepackage{balance}

\volume{Vol. ..., No. ...}
\firstpage{1}
\journalname{VNU Journal of Science: Comp. Science $\&$ Com. Eng.,}

\runauth{Phuong Le-Hong et al.}

\usepackage{amssymb}
\usepackage[figuresright]{rotating}

\usepackage{graphicx}
\usepackage{balance}
\usepackage{color}
\usepackage{tabularx}
\usepackage{array}
\usepackage{rotating}
\usepackage{booktabs}
\usepackage{multirow}
\usepackage{multicol}
\usepackage{amsfonts, amsmath, amssymb}
\usepackage{listings}
\usepackage{epstopdf}
\usepackage{subcaption}
\usepackage{caption}
\usepackage{amsthm}

\usepackage{lipsum}

\usepackage[linesnumbered,ruled,vlined]{algorithm2e}
\usepackage{graphicx}
\usepackage{url}
\usepackage{tikz-qtree}
\usepackage{fixltx2e}
\usepackage{tikz}
\usepackage{tikz-qtree, tikz-qtree-compat}
\usetikzlibrary{fit,shapes}
\usetikzlibrary{plotmarks}
\usepackage{tikz-dependency}
\usepackage{mathtools}
\usepackage{amsmath}

\newproof{pf}{Proof}
\newcommand{\code}[1]{\lstinline|#1|}

\newcommand{\argmax}[1]{\underset{#1}{\operatorname{argmax}}\;}

\SetAlFnt{\scriptsize}
\SetAlCapFnt{\small}

\begin{document}
\captionsenglish
\begin{frontmatter}
\title{Vietnamese Semantic Role Labelling}


\author{Le-Hong Phuong$^{1}$\corref{label1}}
\author{Pham Thai Hoang$^{2}$}
\author{Pham Xuan Khoai$^{2}$}
\author{Nguyen Thi Minh Huyen$^{1}$}
\author{Nguyen Thi Luong$^{3}$}
\author{Nguyen Minh Hiep$^{3}$}


\address{\normalsize $^{1}$Hanoi University of Science, Vietnam National University, Hanoi, Vietnam\\
Email: phuonglh@vnu.edu.vn, huyenntm@vnu.edu.vn\\
$^{2}$FPT University, Hanoi, Vietnam\\
Email: phamthaihoang.hn@gmail.com, phamxuankhoai@gmail.com \\
$^{3}$Dalat University, Lamdong, Vietnam\\
Email:luongnt@dlu.edu.vn, hiepnm@dlu.edu.vn}


\begin{abstract}
In this paper, we study semantic role labelling (SRL), a subtask of
semantic parsing of natural language sentences and its application
for the Vietnamese language. We present our effort in building
Vietnamese PropBank, the first Vietnamese SRL corpus and a software
system for labelling semantic roles of Vietnamese texts. In
particular, we present a novel constituent extraction algorithm in
the argument candidate identification step which is more suitable and
more accurate than the common node-mapping method. In the machine
learning part, our system integrates distributed word features
produced by two recent unsupervised learning models in two learned
statistical classifiers and makes use of integer linear programming
inference procedure to improve the accuracy. The system is evaluated
in a series of experiments and achieves a good result, an $F_1$ score
of 74.77\%. Our system, including corpus and software, is available as
an open source project for free research and we believe that it is a
good baseline for the development of future Vietnamese SRL systems. 
\end{abstract}

\end{frontmatter}

\section{Introduction}
\label{section1}
In this paper, we study semantic role labelling (SRL), a
subtask of semantic parsing of natural language sentences.
SRL is the task of identifying semantic roles of arguments of each predicate in a
sentence. In particular, it answers a question \textit{Who did what to
  whom, when, where, why?}. For each predicate in a sentence, the goal
is to identify all constituents that fill a semantic role, and to
determine their roles, such as agent, patient, or instrument, and
their adjuncts, such as locative, temporal or manner.

Figure~\ref{fig:1} shows the SRL of a simple Vietnamese
sentence. In this example, the arguments of the predicate \textit{giúp} (helped) are labelled with their semantic roles. The meaning of the labels will be described in detail in
Section~\ref{sec:vpropbank}.

\begin{figure}[!h]
\caption{SRL of the Vietnamese sentence "\textit{Nam giúp Huy học bài vào hôm qua}" (Nam helped Huy
to do homework yesterday)}\label{fig:1}
\begin{small}
\begin{equation*}
\underbracket[0.5px]{\text{Nam}}_{\text{Who}} \; \text{giúp} \;
\underbracket[0.5px]{\text{Huy}}_{\text{Whom}} \;
\underbracket[0.5px]{\text{học bài}}_{\text{What}} \;
\underbracket[0.5px]{\text{vào hôm qua}}_{\text{When}} 
\end{equation*}
\begin{equation*}
  \underbracket[0.5px]{\text{Nam}}_{\text{Arg0}} \; \text{giúp}\;
  \underbracket[0.5px]{\text{Huy}}_{\text{Arg1}} \;
  \underbracket[0.5px]{\text{học bài}}_{\text{Arg2}} \;
  \underbracket[0.5px]{\text{vào hôm qua}}_{\text{ArgM-TMP}} 
\end{equation*}
\end{small}
\end{figure}

SRL has been used in many natural language processing (NLP)
applications such as question answering \cite{Shen:2007}, machine
translation \cite{Lo:2010}, document summarization \cite{Aksoy:2009} and information
extraction \cite{Christensen:2010}. Therefore, SRL is an important task in NLP.
The first SRL system was developed by Gildea and
Jurafsky \cite{Gildea:2002}. This system was performed on the English FrameNet
corpus. Since then, SRL task has been widely
studied by the NLP community. In particular, there have been two
shared-tasks, CoNLL-2004 \cite{Carreras:2004} and
CoNLL-2005 \cite{Carreras:2005}, focusing on SRL task for English. Most
of the systems participating in these shared-tasks treated this problem
as a classification problem which can be solved by supervised machine
learning techniques. There exists also several systems for other well-studied languages like Chinese \cite{Xue:2005} or Japanese \cite{Tagami:2009}.

This paper covers not only the contents of two works published in conference proceedings \cite{vnPropBank:2014} (in Vietnamese) and \cite{Pham:2015} on the construction and the evaluation of a first SRL system for Vietnamese, but also an extended investigation of techniques used in SRL. More concretely, the use of integer
linear programming inference procedure and distributed word
representations in our semantic role labelling system, which leads to
improved results over our previous work, as well as a more elaborate
evaluation are new for this article.       

Our system includes two main components, a
SRL corpus and a SRL software which is thoroughly evaluated. 
We employ the same development methodology of the English PropBank
to build a SRL corpus for Vietnamese containing a large number of
syntactically parsed sentences with predicate-argument structures. 

We then use this SRL corpus and supervised machine learning models to
develop a SRL software for Vietnamese. We demonstrate that a simple
application of SRL techniques developed for English or other languages
could not give a good accuracy for Vietnamese. In particular, in the
constituent identification step, the widely used 1-1 node-mapping
algorithm for extracting argument candidates performs poorly on the
Vietnamese dataset, having $F_1$ score of 35.93\%. We thus introduce a
new algorithm for extracting candidates, which is much more accurate,
achieving an $F_1$ score of 84.08\%. In the classification step, in
addition to the common linguistic features, we propose novel and
useful features for use in SRL, including function tags and
distributed word representations. These features are
employed in two statistical classification models, maximum entropy and
support vector machines, which are proved to be good at many
classification problems. In order to incorporate important grammatical
constraints into the system to improve further the performance, we
combine machine learning techniques with an inference procedure based
on integer linear programming. Finally, we use distributed word
representations produced by two recent unsupervised models, the
Skip-gram model and the GloVe model, on a large corpus to alleviate
the data sparseness problem. These word embeddings help our SRL
software system generalize well on unseen words. Our final system
achieves an $F_1$ score of 74.77\% on a test corpus. This system,
including corpus and software, is available as an open source project
for free research and we believe that it is a good baseline for the
development of future Vietnamese SRL systems.

The remainder of this paper is structured as
follows. Section~\ref{sec:corpus} describes the construction of a SRL
corpus for Vietnamese. Section~\ref{sec:software} presents the
development of a SRL software, including the methodologies of existing
systems and of our system. Section~\ref{sec:eval} presents the
evaluation results and discussion. Finally,
Section~\ref{sec:conclusion} concludes the paper and suggests some
directions for future work.

\section{Vietnamese SRL Corpus}
\label{sec:corpus}
Like many other problems in NLP, annotated corpora are essential for statistical learning as well as evaluation of SRL systems. In this section, we start with an introduction of existing English SRL corpora. Then we present our work on the construction of the first reference SRL corpus for Vietnamese.
\subsection{Existing English SRL Corpora}
\label{sec:corpora}
\subsubsection{FrameNet}
The FrameNet project is a lexical database of English. It was built by
annotating examples of how words are used in actual texts. It consists
of more than 10,000 word senses, most of them with annotated examples
that show the meaning and usage and more than 170,000 manually
annotated sentences \cite{Baker:2003}. This is the most widely used
dataset upon which SRL systems for English have been developed and
tested.

\begin{figure}
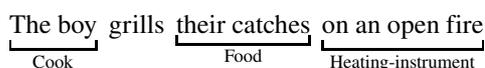

\caption{example sentence in the FrameNet corpus}
\label{fig:4}
\begin{small}
\[
\underbracket[0.5px]{\text{The boy}}_{\text{Cook}} \; \text{grills} \;
\underbracket[0.5px]{\text{their catches}}_{\text{Food}} \;
\underbracket[0.5px]{\text{on an open fire}}_{\text{Heating-instrument}} \;  
\]
\end{small}
\end{figure}

FrameNet is based on the Frame Semantics theory \cite{Boas:2005}. The basic idea is that
the meanings of most words can be best understood on the basis of a
semantic frame: a description of a type of event, relation, or entity
and the participants in it. All members in semantic frames are called
frame elements. For example, a sentence in FrameNet is annotated in
cooking concept as shown in Figure~\ref{fig:4}.

\subsubsection{PropBank}
\begin{figure}
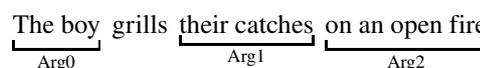

\caption{An example sentence in the PropBank corpus} \label{fig:5}
\begin{small}
\[
\underbracket[0.5px]{\text{The boy}}_{\text{Arg0}} \; \text{grills} \;
\underbracket[0.5px]{\text{their catches}}_{\text{Arg1}} \;
\underbracket[0.5px]{\text{on an open fire}}_{\text{Arg2}} \;  
\]
\end{small}
\end{figure}

PropBank is a corpus that is annotated with verbal propositions and
their arguments \cite{Palmer:2005}. PropBank tries to supply a general purpose labelling
of semantic roles for a large corpus to support the training of
automatic semantic role labelling systems. However, defining such a
universal set of semantic roles for all types of predicates is a
difficult task; therefore, only Arg0 and Arg1 semantic roles can be
generalized. In addition to the core roles, PropBank defines several
adjunct roles that can apply to any verb. It is called Argument
Modifier. The semantic roles covered by the PropBank are the following:
\begin{itemize}
\item Core Arguments (Arg0-Arg5, ArgA): Arguments define
  predicate specific roles. Their semantics depend on predicates in
  the sentence.
\item Adjunct Arguments (ArgM-*): General arguments that can 
  belong to any predicate. There are 13 types of adjuncts.
\item Reference Arguments (R-*): Arguments represent arguments
  realized in other parts of the sentence.
\item Predicate (V): Participant realizing the verb of the
  proposition.
\end{itemize}
For example, the sentence of Figure~\ref{fig:4} can be annotated in 
the PropBank role schema as shown in Figure~\ref{fig:5}.
 
The English PropBank methodology is currently implemented for a wide
variety of languages such as Chinese, Arabic or Hindi with the aim of
creating parallel PropBanks\footnote{\url{http://verbs.colorado.edu/~mpalmer/projects/ace/
EPB-annotation-guidelines.pdf}}. This SRL resource has a great impact on
many natural language processing tasks and applications.

\subsubsection{VerbNet}

VerbNet is a verb lexicon of English, which was developed by Karin
Kipper-Schuler and colleagues \cite{Schuler:2006}. It contains more
than 5800 English verbs, which are classified into 270 groups,
according to the verb classification method of Beth Levin \cite{Levin:1993}.
In this approach, the behavior of a verb is mostly determined by its
meaning. 

Once classified into groups, each verb group is added semantic
roles. VerbNet has 23 semantic roles, for example
\begin{itemize}
\item Actor, the participant that is the investigator of an event.
\item Agent, the actor in an event who initiates and carries out the event
  and who exists independently of the event.
\item Attribute, the undergoer that is a property of an entity or entities.
\item Destination, the goal that is a concrete, physical location.
\end{itemize}
These syntactic roles normally answer who, what, when and how
questions. A SRL annotation guidelines of this project is available online
\footnote{\url{http://verbs.colorado.edu/verb-index/VerbNet_Guidelines.pdf}}. 
In summary, SRL corpora have been constructed for English and other
well-resourced languages. They are important resources
which are very useful for many natural language processing applications. 

For the Vietnamese language, there has not existed any SRL
corpus which with a similar level like those of English corpora
described above. In the following sections, we report our initiatives for
constructing and evaluating a SRL corpus for Vietnamese.

\subsection{Building a Vietnamese PropBank}
\label{sec:vpropbank}

In this section, we present the construction of a Vietnamese SRL
corpus, which is referred as Vietnamese PropBank hereafter. We first
describe annotation guidelines and then describe the SRL corpus which
has been developed.

\subsubsection{Vietnamese SRL Annotation Guidelines}
\label{sec:guidelines}

The determination of semantic roles in the Vietnamese language is a
difficult problem and it has been investigated with different opinions. 
In general, Vietnamese linguists have not reached a consensus on a list of 
semantic roles for the language. Different linguists proposed different lists;
some used the same name but with different meaning of a role, or different names 
having the same meaning. 

Nevertheless, one can use an important principle for determining semantic roles: 
"Semantic role is the actual role a participant plays in some situation and it always 
depends on the nature of that situation" \cite{Cao:2006}. This means that when 
identifying the meaning of a phrase or of a sentence, one must not separate it 
out of the underlying situation that it appears. While there might be
some controversy about the exact semantic role names should be, one
can list common semantic roles which have been accepted by most of
Vietnamese linguists \cite{Hiep:2008}.

The syntactic sub-categorization frames are closely related to the
verb meanings.  That is, the meaning of a sentence can be captured by
the subcategorization frame of the verb predicate. In consequence, the
sentence meaning can be described by labelling the semantic roles for
each participant in the sub-categorization frame of the
predicate. This approach is adopted by many Vietnamese linguists and
different semantic roles set have been proposed. For example,
\textit{Cao Xuân Hạo} \citep{Cao:2006} makes use of the argument
(obligatory participants) roles as agent, actor, processed, force,
carrier, patient, experiencer, goal, etc., while  \textit{Diệp Quang
  Ban} \citep{Diep:1998} makes use of fact categories: dynamic,
static, mental, existential, verbal, relational etc. For adjuncts
(optional participants), \textit{Cao Xuân Hạo} uses the roles: manner,
mean, result, path, etc., while \textit{Diệp Quang Ban} makes use of
circumstance types: time, space, cause, condition, goal, result, path,
etc. 

In this work, we took a pragmatic standpoint during the design of a
semantic role tagset and focused our attention on the SRL categories
that we expect to be most necessary and useful in practical
applications. We have constructed a semantic role tagset based on two
following principles:
\begin{itemize}
\item The semantic roles are well-defined and commonly accepted by the
  Vietnamese linguist community.
\item The semantic roles are comparable to those of the English
  PropBank corpus, which make them helpful and advantageous for
  constructing multi-lingual corpora and applications in later
  steps. Furthermore, it seems fairly indisputable that there are
  structural and semantic correspondences accross languages.
\end{itemize}
We have selected a SRL tagset which is basically similar to that of
the PropBank. However, some roles are made more fine-grained
accounting for idiosyncratic properties of the Vietnamese language. In
addition, some new roles are added to better distinguish predicate
arguments when the predicate is an adjective, a numeral, a noun or a
preposition, which is a common phenomenon in Vietnamese besides the
popular verbal predicate.

\begin{figure}[!h]
\tikzset{
  sibling distance=-3pt
}
\centering
\caption{A SRL annotated Vietnamese sentence}
\label{fig:example4}
\begin{small}
\begin{tikzpicture}
\Tree
[.S
    \edge node[auto=right] {\texttt{Arg0}};
    [.NP-SUB 
       [.M Ba ]
       [.Nc-H đứa ]
       [.N con ]
       [.N anh ]
        ]
    [.VP 
        \edge node[midway,left] {\texttt{Arg-TMP}};
        [.R đã ]
        \edge node[midway,right] {\texttt{Rel}};
        [.V-H có ]
        \edge node[midway,right] {\texttt{Arg1}};
        [.NP-DOB 
        	[.N-H {việc làm} ]
        	[.A {ổn định} ]
         ]
        ]
]
\end{tikzpicture}
\end{small}
\end{figure}
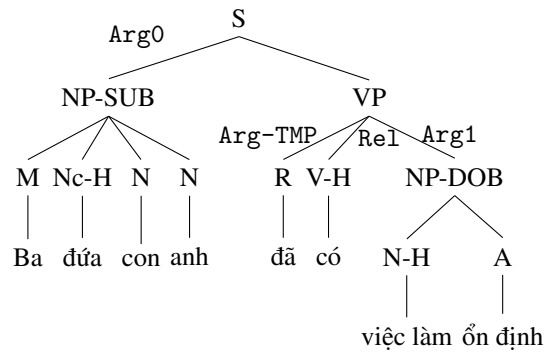

The following paragraph describes some semantic roles of predicative
arguments where the predicate is a verb:
\begin{itemize}
\item \texttt{Arg0}: The agent semantic role representing a 
  person or thing who is the doer of an event. For example,
  \begin{equation*}
    \underbracket[0.5px]{\text{Nam}}_{\mathtt{Arg0}} \text{ đến trường} \; (\text{Nam goes to school.})
  \end{equation*}
\item \texttt{Arg0-Identified} and \texttt{Arg1-Identifier}: The
  semantic roles representing identified entity and identifier
  respectively, normally used with the copula ``là''. For example,
  \begin{equation*}
  \underbracket[0.5px]{\text{Cầu thủ giỏi nhất ở
      đây}}_{\mathtt{Arg0-Identified}} \text{ là }
  \underbracket[0.5px]{\text{anh ấy}}_{\mathtt{Arg1-Identifier}}   
  \end{equation*}
  $
  (\text{He is the best player here.})$
\item \texttt{Arg1-Patient}: The semantic role which is the surface
  object of a predicate indicating the person or thing affected. For
  example, 
  \begin{equation*}
    \text{Bộ đội phá }
      \underbracket[0.5px]{\text{cầu}}_{\texttt{Arg1-Patient}} \;     
  \end{equation*}
  $$(\text{The soldiers broke a bridge.})$$
\item \texttt{Arg2}: The semantic role of a beneficiary indicating a
  referent who is advantaged or disadvantaged by an event. For
  example,
  \begin{equation*}
    \text{Nó chữa cái xe cho } \underbracket[0.5px]{\text{chị
        ấy}}_{\mathtt{Arg2}} \; 
  \end{equation*}
  $$(\text{He repaired a bike for her.})$$
\end{itemize}
Figure~\ref{fig:example4} presents an example of the SRL analysis of a
syntactically bracketed sentence ``\textit{Ba đứa con anh đã có việc làm ổn
định}.'' (His three children have had a permanent job.). The semantic
roles of this sentence include:
\begin{itemize}
\item \texttt{Arg0}: ``\textit{ba đứa con anh}'' (his three children) is the agent
\item \texttt{ArgM-TMP}: ``\textit{đã}'' is a temporal modifier
\item \texttt{Rel}: ``\textit{có}'' (have) is the predicate
\item \texttt{Arg1}: ``\textit{việc làm ổn định}'' (a permanent job) is the patient.
\end{itemize}

\subsubsection{Vietnamese SRL Corpus}

Once the SRL annotation guidelines have been designed, we built a
Vietnamese SRL corpus by following two main steps.

In the first step, we proposed a set of conversion rules to convert automatically a syntactically
annotated treebank containing 10,000 manually annotated sentences
(the VietTreeBank) to a coarse-grained SRL annotated corpus. 

The Vietnamese treebank is one result of a national project which aims
to develop basic resources and tools for Vietnamese language and
speech processing\footnote{VLSP Project,
  \url{https://vlsp.hpda.vn/demo/}}. The raw texts of the
treebank are collected from the social and political sections of the
Youth online daily newspaper. The corpus is divided into three sets
corresponding to three annotation levels: word-segmented,
part-of-speech-tagged and syntax-annotated set. The syntax-annotated
corpus, a subset of the part-of-speech-tagged set, is currently
composed of $10,471$ sentences ($225,085$ tokens).  Sentences range
from $2$ to $105$ words, with an average length of $21.75$
words. There are $9,314$ sentences of length $40$ words or less. The
tagset of the treebank has $38$ syntactic labels ($18$ part-of-speech
tags, $17$ syntactic category tags, $3$ empty categories) and 17
function tags. For details, please refer
to \cite{VTB:2009}\footnote{All the resources are available at the
  website of the VLSP project.}. The meanings of some common tags are
listed in Table~\ref{tab:vtb}.

\begin{table}[!h]
\centering
  \caption{Some Vietnamese treebank tags}\label{tab:vtb}
  \begin{small}
  \begin{tabular}{ r  c  l }
    \hline\hline
    No.&Category&Description\\
    \hline
    1.&S&simple declarative clause\\
    2.&VP&verb phrase\\
    3.&NP&noun phrase\\
    4.&PP&preposition phrase\\
    5.&N&common noun\\
    6.&V&verb\\
    7.&P&pronoun\\
    8.&R&adverb\\
    9.&E&preposition\\
    10.&CC&coordinating conjunction\\
    \hline\hline
  \end{tabular}
  \end{small}
\end{table}

The coarse-grained semantic role tagset contains 24 role names which
are all based on the main roles of the PropBank. We carefully investigated the
tagset of the VietTreeBank based on detailed guidelines of constituency
structures, phrasal types, functional tags, clauses, parts-of-speech
and adverbial functional tagset to propose a set of rules for
determining high-level semantic roles. Some rules for coarse-grained
annotation are shown in Table~\ref{tab:conversion-rules}. Each rule is
used to determine a semantic role for a phrase of a sentence.

As an example, consider the constituency analysis of a sentence in the
VietTreeBank ``\textit{Kia là những ngôi nhà vách đất.}'' (Over there
are soil-wall houses.)
\begin{verbatim}
(S (NP-SUB (P-H Kia)) (VP (V-H là) 
   (NP (L những) (Nc-H ngôi) (N nhà) 
      (NP (N-H vách) (N đất)))) (. .))
\end{verbatim}
First, using the annotation rule for \texttt{Arg0}, the phrase having
syntactical function \texttt{SUB} or preceding the predicate of the
sentence, we can annotate the semantic role \texttt{Arg0} for the word
``\textit{Kia}''. The predicate ``\textit{là}'' is annotated with semantic
role \texttt{REL}. Finally, the noun phrase following the predicate
``\textit{những ngôi nhà vách đất}'' is annotated with \texttt{Arg1}.

\begin{table}[!h]
\centering
\caption{Some rules for coarse-grained SRL annotation}\label{tab:conversion-rules}
\begin{small}
\begin{tabular}{l l l}
\hline\hline
 Role& Description & Rule\\
\hline
\texttt{ARG0} & Agent & SUB | Phrasal types \\
 & & (NP, ...) preceding \\
  & & predicate\\
\texttt{ARG1} & Patient & DOB | phrasal types\\
 & & (NP, ...) following\\
  & & predicate\\
\texttt{ARG2} & Beneficiary & IOB phrases\\
\texttt{ARGM-NEG} & Negation & Negative words \\
 &  & ''\textit{không, chẳng, }\\
  &  & \textit{chớ, chả}\dots''\\
\texttt{ARGM-LOC} & Locatives & LOC phrases\\
\texttt{ARGM-MNR} & Manner & MNR phrases\\
 & markers & \\
\texttt{ARGM-CAU} & Cause & PRP | causal words \\
 & clauses &``\textit{do, bởi vì, }\\
  & clauses &\textit{vì, bởi},\dots''\\
\texttt{ARGM-DIR} & Directionals & DIR phrases\\
\texttt{ARGM-DIS} & Conjunctive & CC phrases \\
 & clauses & or C word\\
\texttt{ARGM-EXT} & Extent markers & EXT phrases\\
\dots && \\
\hline\hline
\end{tabular}
\end{small}
\end{table}

In the second step, we developed a software to help a team of
Vietnamese linguists manually revise and annotate the converted corpus with
fine-grained semantic roles. The software is web-based, friendly and
easy for correction and edition of multiple linguists. In addition, it
also permits a collaborative work where any edition at sentence level
is versionized and logged with meta-information so as to facilitate
cross validation and discussion between linguists if necessary.

We have completed the semantic role annotation of 5,460 sentences of
the VietTreeBank, covering 7,525 verbal and adjectival
predicatives. The annotation guidelines as well as the current SRL
corpus are published as open resources for free research.

In the next section, we present our effort in developing a SRL
software system for Vietnamese which is constructed and evaluated on this
SRL corpus. 

\section{Vietnamese SRL System}
\label{sec:software}

\subsection{Existing Approaches}
\label{sec:approaches}

This section gives a brief survey of common approaches which are used
by many existing SRL systems of well-studied languages. These systems
are investigated in two aspects: (a) the data type that the systems
use and (b) their approaches for labelling semantic roles, including
model types, labelling strategies, degrees of granularity and
post-processing.

\subsubsection{Data Types}
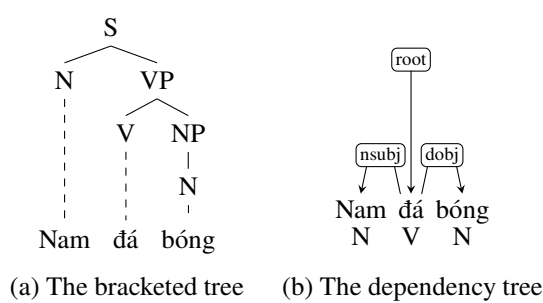
\begin{figure}[!h]
\centering
\caption{Bracketed and dependency trees for sentence \textit{Nam đá
    bóng} (Nam plays football)} \label{fig:6}
\begin{small}	
\begin{tabular}{cc}
\begin{tikzpicture}
\tikzset {level distance=20pt}
\tikzset {frontier/.style={distance from root=80pt}}
\Tree [.S [.N \edge[dashed]; Nam ]
		  [.VP [.V \edge[dashed]; đá ]
		  	   [.NP [.N \edge[dashed]; bóng ] ] ] ]
\end{tikzpicture} &
\begin{dependency}
\begin{deptext}
Nam \& đá \& bóng \\
N   \& V  \& N \\
\end{deptext}
\deproot{2}{root}
\depedge{2}{1}{nsubj}
\depedge{2}{3}{dobj}
\end{dependency} \\
(a) {The bracketed tree} & (b) {The dependency tree}
\end{tabular}
\end{small}
\end{figure}

The input data of a SRL system are typically syntactically parsed
sentences. There are two common syntactic representations namely
bracketed trees and dependency trees. 
Some systems use bracketed trees of sentences as input data.  A bracketed tree of a
sentence is the tree of nested constituents representing its
constituency structure. Some systems use dependency trees of a
sentence, which represents dependencies between individual words of a
sentence.  The syntactic dependency represents the fact that the
presence of a word is licensed by another word which is its
governor. In a typed dependency analysis, grammatical labels are added
to the dependencies to mark their grammatical relations, for example
\textit{nominal subject} (nsubj) or \textit{direct object}
(dobj). Figure~\ref{fig:6} shows the bracketed tree and the
dependency tree of an example sentence.

\subsubsection{SRL Strategy}

\paragraph{Input Structures}

The first step of a SRL system is to extract constituents
that are more likely to be arguments or parts of arguments. This step
is called argument candidate extraction. Most of SRL systems for
English use 1-1 node mapping method to find candidates. This method searches all nodes in a parse tree and
maps constituents and arguments. Many systems use a pruning strategy
on bracketed trees to better identify argument
candidates \cite{Xue:2005}. 

\paragraph{Model Types}
In a second step, each argument candidate is labelled 
with a semantic role. Every SRL system has a classification model
which can be classified into two types, independent model or joint
model. While an independent model decides the label of each
argument candidate independently of other candidates, a joint model
finds the best overall labelling for all candidates in the
sentence at the same time. Independent models are fast but are prone to
inconsistencies such as argument overlap, argument repetition or 
argument missing.  For example, Figure~\ref{fig:7} 
shows some examples of these inconsistencies when analyzing the 
Vietnamese sentence \textit{Do học chăm, Nam đã đạt
  thành tích cao} (By studying hard, Nam got a high achievement). 

\begin{figure}[!h]
\centering
\caption{Examples of some inconsistencies}\label{fig:7}
\begin{small}
\begin{subfigure}{0.33 \textheight}
\centering
\[
\text{Do}\underbracket{\text{học chăm}}_{Arg1}, \text{ Nam đã \textit{đạt} thành tích cao.}
\]
\[
\text{Do học} \underbracket{\text{chăm, Nam}}_{Arg1}\text{ đã \textit{đạt} thành tích cao.}
\]
\caption{Overlapping argument}
\end{subfigure}
\begin{subfigure}{0.33 \textheight}
\centering
\[
\text{Do} \underbracket{\text{học}}_{Arg1} \text{chăm}, \underbracket{\text{Nam}}_{Arg1} \text{đã \textit{đạt} thành tích cao.}
\]
\caption{Repeated argument}
\end{subfigure}
\begin{subfigure}{0.33 \textheight}
\centering
\[
\underbracket{\text{Do học chăm, Nam}}_{Arg0} \text{đã \textit{đạt}} \underbracket{\text{thành tích cao}}_{Arg0}.
\]
\caption{Missing argument}
\end{subfigure}
\end{small}
\end{figure}

\paragraph{Labelling Strategies}
Strategies for labelling semantic roles are diverse, but they can be
classified into three main strategies. Most of the systems use a
two-step approach consisting of identification and
classification \cite{Punyakanok:2005,Haghighi:2005}. The first step
identifies arguments from many candidates, which is essentially a
binary classification problem. The second step classifies the
identified arguments into particular semantic roles. Some systems use
a single classification step by adding a ``null'' label into semantic
roles, denoting that this is not an
argument \cite{Surdeanu:2005}. Other systems consider SRL as a
sequence tagging problem \cite{Marquez:2005,Pradhan:2005}.

\paragraph{Granularity}
Existing SRL systems use different degrees of granularity when
considering constituents. Some systems use individual words as their
input and perform sequence tagging to identify arguments. This method
is called word-by-word (W-by-W) approach. Other systems use syntactic
phrases as input constituents. This method is called
constituent-by-constituent (C-by-C) approach. Compared to the W-by-W
approach, C-by-C approach has two main advantages. First, phrase
boundaries are usually consistent with argument boundaries. Second,
C-by-C approach allows us to work with larger contexts due to a
smaller number of candidates in comparison to the W-by-W
approach. Figure~\ref{fig:gran} presents an example of C-by-C and
W-by-W approaches.

\begin{figure}[!h]
\begin{small}
\centering
\caption{C-by-C and W-by-W approaches} \label{fig:gran}
\begin{subfigure}{0.3\textheight}
\centering
\[
\underbracket[0.5px]{\text{Nam}} \; gi\acute{u}p \;
\underbracket[0.5px]{\text{Huy}} \; \underbracket[0.5px]{\text{học bài}} \; \underbracket[0.5px]{\text{v\`{a}o hôm qua}} 
\]
\caption{Example of C-by-C}
\end{subfigure}
\begin{subfigure}{0.3\textheight}
\centering
\[
\underbracket[0.5px]{\text{Nam}} \; gi\acute{u}p \;
\underbracket[0.5px]{\text{Huy}} \; \underbracket[0.5px]{\text{học}} \; \underbracket[0.5px]{\text{bài}} \; \underbracket[0.5px]{\text{v\`{a}o}} \; \underbracket[0.5px]{\text{hôm qua}} 
\]
\caption{Example of W-by-W}
\end{subfigure}
\end{small}
\end{figure}

\paragraph{Post-processing}
To improve the final result, some systems use post-processing to
correct argument labels. Common post-processing methods include 
re-ranking, Viterbi search and integer linear programming (ILP).

\subsection{Our Approach}
\label{sec:our-approach}

The previous subsection has reviewed existing techniques for SRL which
have been published so far for well-studied languages. In this
section, we first show that these techniques per se cannot give a good
result for Vietnamese SRL, due to some inherent difficulties, both in
terms of language characteristics and of the available corpus. We then
develop a new algorithm for extracting candidate constituents for use
in the identification step.

Some difficulties of Vietnamese SRL are related to its SRL corpus. As
presented in the previous section, this SRL corpus has 5,460
annotated sentences, which is much smaller than SRL corpora of other
languages. For example, the English PropBank contains about 50,000
sentences, which is about ten times larger. While smaller in size, the
Vietnamese PropBank has more semantic roles than the English PropBank
has -- 28 roles compared to 21 roles. This makes the unavoidable data
sparseness problem more severe for Vietnamese SRL than for English
SRL.

In addition, our extensive inspection and experiments on the Vietnamese
PropBank have uncovered that this corpus has many annotation errors,
largely due to encoding problems and inconsistencies in annotation. In
many cases, we have to fix these annotation errors by ourselves. In
other cases where only a proposition of a complex sentence is
incorrectly annotated, we perform an automatic preprocessing procedure
to drop it out, leave the correctly annotated propositions
untouched. We finally come up with a corpus of 4,800 sentences which
are semantic role annotated. 


A major difficulty of Vietnamese SRL is due to the nature of the
language, where its linguistic characteristics are different from
occidental languages \cite{Le-Hong:2015}. We first try to apply the
common node-mapping algorithm which is widely used in English SRL
systems to the Vietnamese corpus. However, this application gives us a
very poor performance. Therefore, in the identification step, we
develop a new algorithm for extracting candidate constituents which is
much more accurate for Vietnamese than the node-mapping
algorithm. Details of experimental results will be provided in the
Section~\ref{sec:eval}.

In order to improve the accuracy of the classification step,
and hence of our SRL system as a whole, we have integrated many useful
features for use in two statistical classification models, namely
Maximum Entropy (ME) and Support Vector Machines (SVM). On the one
hand, we adapt the features which have been proved to be good for SRL
of English. On the other hand, we propose some novel features,
including function tags, predicate type and distance. Moreover, to
improve further the performance of our system, we introduce some
appropriate constraints and apply a post-processing
method by using ILP. Finally, to better handle unseen words, we generalize 
the system by integrating distributed word representations.

In the next paragraphs, we first present our constituent extraction
algorithm to get inputs for the identification step and then the ILP
post-processing method. Details of the features used in the
classification step and the effect of distributed word representations
in SRL will be presented in Section~\ref{sec:eval}.

\subsubsection{Constituent Extraction Algorithm}

Our algorithm derives from the pruning algorithm for
English \cite{Xue:2004} with some modifications. While the original
algorithm collects sisters of the current node, our algorithm checks
the condition whether or not children of each sister have the same
phrase label and have different function label from their parent. If they
have the same phrase labels and different function labels from their
parent, our algorithm collects each of them as an argument
candidate. Otherwise, their parent is collected as a candidate. In
addition, we remove the constraint that does not collect coordinated
nodes from the original algorithm. 

This algorithm aims to extract constituents from a bracketed
tree which are associated to their corresponding predicates of 
the sentence. If the sentence has multiple predicates, multiple
constituent sets corresponding to the predicates are extracted. The pseudo
code of the algorithm is described in Algorithm~\ref{alg:1}. 

\begin{algorithm}[!h]
\DontPrintSemicolon
\KwData{A bracketed tree $T$ and its predicate}
\KwResult{A tree with constituents for the predicate}


\Begin{
$currentNode \leftarrow predicateNode$\;
\While{$currentNode \neq$ T.root()}{
	\For{$S \in currentNode$.sibling()}{
		\If{$|S.children()| > 1$ and $S.children().get(0).isPhrase()$}{
			$sameType \leftarrow true$\\
			$diffTag \leftarrow true$\\
			$phraseType \leftarrow S.children().get(0).phraseType()$\\
			$funcTag \leftarrow S.children().get(0).functionTag()$\\
			\For{$i\leftarrow 1$ \KwTo $|S.children()| - 1$}{
				\If{$S.children().get(i).phraseType() \neq phraseType$}{
					$sameType \leftarrow false$\\
					break
				}
				\If{$S.children().get(i).functionTag() = funcTag$}{
					$diffTag \leftarrow false$\\
					break
				}
			}
			\If{$sameType$ and $diffTag$}{
				\For{$child \in S.children()$}{
					$T.collect(child)$
				}
			}
		}
		\Else{
			$T.collect(S)$
		}
	}
	$currentNode \leftarrow currentNode.parent()$
}
return $T$
}
\caption{\small Constituent Extraction Algorithm}\label{alg:1}
\end{algorithm}

This algorithm uses several simple functions. The $root()$ function gets the root
of a tree. The $children()$ function gets the children
of a node. The $sibling()$ function gets the sisters of a node. The $isPhrase()$
function checks whether a node is of phrasal type or not. The
$phraseType()$ function and $functionTag()$ function extracts the phrase type 
and function tag of a node, respectively. Finally, the
$collect(node)$ function collects words from leaves of the subtree
rooted at a node and creates a constituent.

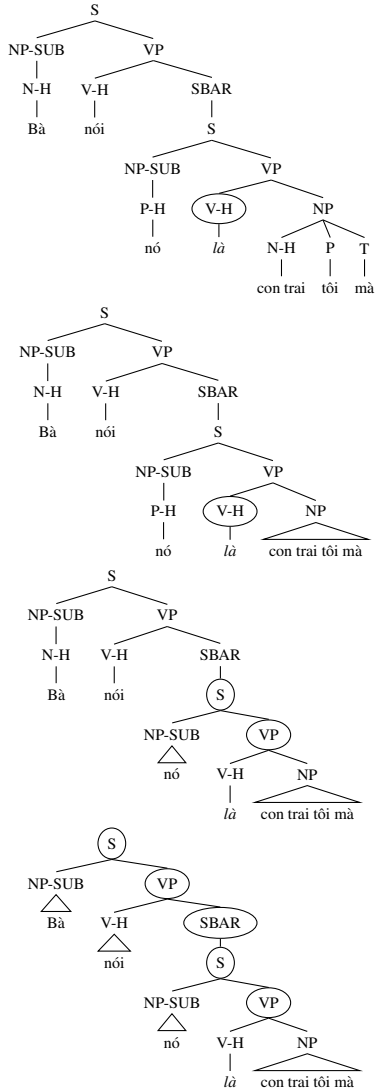
\begin{figure}[!h]
\caption{Extracting constituents of the sentence "Bà nói nó
    là con trai tôi mà" at predicate "\textit{là}"}
\label{fig:8}
\begin{tiny}
\centering
\begin{subfigure}{0.35 \textheight}
\centering
\tikzset {level distance=15pt}
\begin{tikzpicture}
\Tree
[.S [.NP-SUB [.N-H Bà ] ]
	[.VP [.V-H nói ]
		 [.SBAR [.S [.NP-SUB [.P-H nó ] ]
		  	   	    [.VP [.\node[draw,ellipse]{V-H}; \textit{là} ]
		  	   			 [.NP [.N-H {con trai} ]
		  	   			 	  [.P tôi ]
		  	   			 	  [.T mà ] ] ] ] ] ] ] ]
\end{tikzpicture}
\end{subfigure}
\begin{subfigure}{0.35 \textheight}
\centering
\tikzset {level distance=15pt}
\begin{tikzpicture}
\Tree
[.S [.NP-SUB [.N-H Bà ] ]
	[.VP [.V-H nói ]
		 [.SBAR [.S [.NP-SUB [.P-H nó ] ]
		  	   	    [.VP [.\node[draw,ellipse]{V-H}; \textit{là} ]
		  	   			 [.NP \edge[roof]; {con trai tôi mà } ] ] ] ] ] ]
\end{tikzpicture}
\end{subfigure}
\begin{subfigure}{0.35 \textheight}
\centering
\tikzset {level distance=15pt}
\begin{tikzpicture}
\Tree
[.S [.NP-SUB [.N-H Bà ] ]
	[.VP [.V-H nói ]
		 [.SBAR [.\node[draw, ellipse]{S}; [.NP-SUB \edge[roof]; {nó} ]
		  	   	    [.\node[draw, ellipse]{VP}; [.V-H \textit{là} ]
		  	   			 [.NP \edge[roof]; {con trai tôi mà } ] ] ] ] ] ]
\end{tikzpicture}
\end{subfigure}
\begin{subfigure}{0.35 \textheight}
\centering
\tikzset {level distance=15pt}
\begin{tikzpicture}
\Tree
[.\node[draw, ellipse]{S}; [.NP-SUB \edge[roof]; {Bà} ]
	[.\node[draw, ellipse]{VP}; [.V-H \edge[roof]; nói ]
		 [.\node[draw, ellipse]{SBAR}; [.\node[draw, ellipse]{S}; [.NP-SUB \edge[roof]; {nó} ]
		  	   	    [.\node[draw, ellipse]{VP}; [.V-H \textit{là} ]
		  	   			 [.NP \edge[roof]; {con trai tôi mà } ] ] ] ] ] ]
\end{tikzpicture}
\end{subfigure}
\end{tiny}
\end{figure}

Figure~\ref{fig:8} shows an example of running the algorithm on a
sentence \textit{Bà nói nó là con trai tôi mà} (You said that he is my
son). First, we find the current predicate node V-H \textit{là} (is). The current
node has only one sibling NP node. This NP node has three children where
some of them have different labels from their parents, so this node
and its associated words are collected. After that, we set current
node to its parent and repeat the process until reaching the root of the
tree. Finally, we obtain a tree with the following constituents for predicate
\textit{là}: \textit{Bà}, \textit{nói}, \textit{nó}, and \textit{con
  trai tôi mà}.

\subsubsection{Integer Linear Programming}

Because the system classifies arguments independently, labels assigned
to arguments in a sentence may violate Vietnamese grammatical
constraints. To prevent such violation and improve the result, we
propose a post-processing process which finds the best global
assignment that also satisfies grammatical constraints. Our work is based on
the ILP method of English PropBank \cite{Punyakanok:2004}. 
Some constraints that are unique to Vietnamese are also introduced and
incorporated. 

Integer programs are almost identical to linear programs. The cost
function and the constraints are all in linear form. The only difference
is that the variables in ILP can only take integer values. A general
binary ILP can be stated as follows.

Given a cost vector $\vec{p} \in \mathbb R^{d}$, a set of variables
$\vec z = (z_1,\dots, z_d) \in \mathbb R^d$, and cost matrices
$\mathbf{C}_{1} \in \mathbb R^{t_{1}}\times 
\mathbb R^{d}$, $\mathbf{C}_{2} \in \mathbb R^{t_{2}}\times \mathbb
R^{d}$, where $t_1, t_2$ are the number of inequality and equality
constraints and $d$ is the number of binary variables. The ILP
solution $\hat{\vec{z}}$ is the vector that maximizes the cost function:
\begin{equation}
\hat{\vec{z}} = \argmax{\vec{z} \in 0,1^{d}} \vec{p} \cdot  \vec{z}
\quad \text{ subject  to } 
\begin{cases}
\mathbf{C}_{1}\vec{z} \geq \vec{b}_{1}\\
\mathbf{C}_{2}\vec{z} = \vec{b}_{2}
\end{cases}
\end{equation}
where $\vec{b}_1, \vec{b}_2 \in \mathbb R^d$.

Our system attempts to find exact roles for argument
candidate set for each sentence. This set is denoted as $S^{1:M}$,
where the index ranged from $1$ to $M$; and the argument role set is
denoted as $\mathcal P$. Assuming that the classifier returns a score,
$score(S^{i}=c^{i})$, corresponding to the likelihood of assigning
label $c^{i}$ to argument $S^{i}$. The aim of the system is to find 
the maximal overall score of the arguments:
\begin{equation}
\hat{c}^{1:M} = \argmax{c^{1:M} \in \mathcal P^M} score(S^{1:M} = c^{1:M})
\end{equation}
\begin{equation}
= \argmax{c^{1:M} \in \mathcal P^M} \sum^{M}_{i=1} score(S^{i}=c^{i})
\end{equation}

\paragraph{ILP Constraints}
In this paragraph, we propose a constraint set
for our SRL system. Some of them are directly inspired and derived from
results for English SRL, others are constraints that we specify
uniquely to account for Vietnamese specificities. The constraint set includes:
\begin{enumerate}
\item One argument can take only one type.
\item Arguments cannot overlap with the predicate in the sentence.
\item Arguments cannot overlap other arguments in the sentence.
\item There is no duplicating argument phenomenon for core arguments in the sentence.
\item If the predicate is not verb type, there are only 2 types of
  core argument \texttt{Arg0} and \texttt{Arg1}.
\end{enumerate}
In particular, constraints from 1 to 4 are derived from the ILP method
  for English \cite{Punyakanok:2004}, while constraint 5 is designed
  specifically for Vietnamese.
 
\paragraph{ILP Formulation}
To find the best overall labelling satisfying these constraints,
we transform our system to an ILP problem. First, let $z_{ic} =
[S^{i} = c]$ be the binary variable that shows whether or not $S^{i}$ is
labelled argument type $c$. We denote $p_{ic} =
score(S^{i}=c)$. The objective function of the optimization problem can be written as: 
\begin{equation}
\argmax{z \in {0,1}} \sum_{i=1}^{M}\sum_{c=1}^{|\mathcal P|}p_{ic}z_{ic}.
\end{equation}
Next, each constraint proposed above can be reformulated as follows:
\begin{enumerate}
\item One argument can take only one type.
\begin{equation}
\sum_{c=1}^{|\mathcal P|}z_{ic}=1, \quad \forall i \in [1,M].
\end{equation}
\item Arguments cannot overlap with the predicate in the sentence.
\item Arguments cannot overlap other arguments in the sentence.
  If there are $k$ arguments $S^{1},S^{2},...,S^{k}$ that appear in a
  same word in the sentence, we can conclude that there are at least
  $k-1$ arguments that are classified as ``null'':
  \begin{equation}
    \sum_{i=1}^{k}z_{ic} \geq k-1 \quad (c = \text{``null''}).
  \end{equation}
  This constraint has been satisfied by our constituent extraction
  approach. Thus, we do not need to add this constraint in the 
  post-processing step if the constituent extraction algorithm has been
  used. 
\item There is no duplicating argument phenomenon for core arguments in the sentence.
  \begin{equation}
  \begin{split}
    &\sum_{i=1}^{M}z_{ic} \leq 1,  \\
	&\forall c \in \left
      \{\text{Arg0}, \text{Arg1}, \text{Arg2}, \text{Arg3},
      \text{Arg4} \right \}. 
	\end{split}
  \end{equation}
\item If the predicate is not verb type, there are only 2 types of core argument \texttt{Arg0} and \texttt{Arg1}.
  \begin{equation}
    \sum_{i=1}^{M}z_{ic}=0 \quad \forall c \in \left \{\text{Arg2}, \text{Arg3}, \text{Arg4} \right \}.
  \end{equation}
\end{enumerate}

In the next section, we present experimental results, system
evaluation and discussions.

\section{Evaluation}
\label{sec:eval}

In this section, we describe the evaluation of our SRL system. First, we
first introduce two feature sets used in machine learning
classifiers. Then, the evaluation results are presented and
discussed. Next, we report the improved results by using integer
linear programming inference method. Finally, we present the efficacy
of distributed word representations in generalizing the system to
unseen words. 

\subsection{Feature Sets}
\label{sec:features}

We use two feature sets in this study. The first one is composed of
basic features which are commonly used in SRL system for English. This
feature set is used in the SRL system of Gildea and Jurafsky
\citep{Gildea:2002} on the FrameNet corpus.

\subsubsection{Basic Features}

This feature set consists of 6 feature templates, as follows:
\begin{enumerate}
\item Phrase type: This is very useful feature in classifying semantic roles
because different roles tend to have different syntactic
categories. For example, in the sentence in Figure~\ref{fig:8} \textit{Bà nói nó
  là con trai tôi mà}, the phrase type of constituent \textit{nó} is
\textit{NP}. 
\item Parse tree path: This feature captures the syntactic relation between a
constituent and a predicate in a bracketed tree. This is the
shortest path from a constituent node to a predicate node in the
tree. We use either symbol $\uparrow$ or symbol $\downarrow$ to indicate the upward
direction or the downward direction, respectively. For example, the
parse tree path from constituent \textit{nó} to the predicate
\textit{l\`{a}} is \textit{NP$\uparrow$S$\downarrow$VP$\downarrow$V}.
\item Position: Position is a binary feature that describes whether
  the constituent occurs after or before the predicate. 
  It takes value \textit{0} if the constituent appears
  before the predicate in the sentence or value \textit{1}
  otherwise. For example, the position of constituent \textit{nó} in
  Figure~\ref{fig:8} is \textit{0} since it appears before predicate
  \textit{là}.
\item Voice: Sometimes, the differentiation between active and passive voice is
useful. For example, in an active sentence, the subject is
usually an \textit{Arg0} while in a passive sentence, it is often an
\textit{Arg1}. Voice feature is also binary feature, taking value \textit{1} for
active voice or \textit{0} for passive voice. The sentence in
Figure~\ref{fig:8} is of active voice, thus its voice feature value is \textit{1}.
\item Head word: This is the first word of a phrase. For
  example, the head word for the phrase \textit{con trai tôi mà} is
  \textit{con trai}. 
\item Subcategorization: Subcategorization feature captures the tree that
has the concerned predicate as its child. For example, in
Figure~\ref{fig:8}, the subcategorization of the predicate
\textit{l\`{a}} is \textit{VP(V, NP)}.
\end{enumerate}

\subsubsection{New Features}

Preliminary investigations on the basic feature set give us a rather poor
result. Therefore, we propose some novel
features so as to improve the accuracy of the system. These 
features are as follows:
\begin{enumerate}
\item  Function tag: Function tag is a useful information,
especially for classifying adjunct arguments. It determines a
constituent's role, for example,  the function tag of constituent
\textit{nó} is \textit{SUB}, indicating that this has a subjective role.
\item Distance: This feature records the length of the full
  parse tree path before pruning. For example, the distance from constituent
\textit{nó} to the predicate \textit{là} is \textit{3}. 
\item Predicate type: Unlike in English, the type of predicates
  in Vietnamese is much more complicated. It is not only a verb, but is also a noun, an
  adjective, or a preposition. Therefore, we propose
a new feature which captures predicate types. For example, the
predicate type of the concerned predicate is \textit{V}. 
\end{enumerate}

\subsection{Results and Discussions}
\label{sec:results}

\subsubsection{Evaluation Method}

We use a 10-fold cross-validation method to evaluate our system. The
final accuracy scores is the average scores of the 10 runs.

The evaluation metrics are the precision, recall and $F_1$-measure. The
precision ($P$) is the proportion of labelled arguments identified by
the system which are correct; the recall ($R$) is the proportion of
labelled arguments in the gold results which are correctly identified
by the system; and the $F_1$-measure is the harmonic mean of $P$ and
$R$, that is $F_{1} = 2PR/(P+R)$.

\subsubsection{Baseline System}

In the first experiment, we compare our constituent extraction
algorithm to the 1-1 node mapping and the pruning algorithm \cite{Punyakanok:2004}. Table~\ref{tab:per1}
shows the performance of two extraction algorithms.

\begin{table}[!h]
\tikzset{
  sibling distance=-8pt
}
\centering
\caption{Accuracy of three extraction algorithms}\label{tab:per1}
\begin{small}
\begin{tabular}{lccc}
\hline\hline
 & 1-1 Node & Pruning & Our \\
  & Mapping & Alg. & Extraction \\
  & Alg. & & Alg. \\
\hline
Precision & 29.58\% & 85.05\% & 82.15\% \\ 
Recall & 45.82\% & 79.39\% & 86.12\% \\ 
$F_1$ & 35.93\% & 82.12\% & \textbf{84.08\%} \\ 
\hline\hline
\end{tabular}
\end{small} 
\end{table}

We see that our extraction algorithm outperforms significantly the 1-1
node mapping algorithm, in both of the precision and the recall
ratios. It is also better than the pruning algorithm.
In particular, the precision of the 1-1 node mapping algorithm 
is only 29.58\%; it means that this method captures many candidates
which are not arguments. In contrast, our algorithm is able to
identify a large number of correct argument candidates, particularly
with the recall ratio of 86.12\% compared to 79.39\% of the pruning
algorithm. This result also shows that we cannot take for
granted that a good algorithm for English could also work well for
another language of different characteristics. 

In the second experiment, we continue to compare the performance of
the two extraction algorithms, this time at the final classification
step and get the baseline for Vietnamese SRL. The classifier we use in
this experiment is a Support Vector Machine (SVM)
classifier\footnote{We use the linear SVM classifier with $L_2$
  regularization provided by the \texttt{scikit-learn} software
  package. The regularization term is fixed at 0.1.}.
Table~\ref{tab:per2} shows the accuracy of the baseline system.

\begin{table}[!h]
\centering
\caption{Accuracy of the baseline system} \label{tab:per2}
\begin{small}
\begin{tabular}{lccc}
\hline\hline
 & 1-1 Node & Pruning & Our  \\
 & Mapping & Alg. & Extraction \\ 
  & Alg. &  & Alg. \\ 
\hline
Precision & 66.19\% & 73.63\% & 73.02\% \\ 
Recall & 29.34\% & 62.79\% & 67.16\% \\ 
$F_1$ & 40.66\% & 67.78\% & \textbf{69.96\%} \\ 
\hline\hline
\end{tabular}
\end{small} 
\end{table}

Once again, this result confirms that our algorithm achieves the
better result. The $F_1$ of our baseline SRL system is 69.96\%,
compared to 40.66\% of the 1-1 node mapping and 67.78\% of the pruning
system. This result can be explained by the fact that the 1-1 node
mapping and the pruning algorithm have a low recall ratio, because it
identifies incorrectly many argument candidates.

\subsubsection{Labelling Strategy}

In the third experiment, we compare two labelling strategies for
Vietnamese SRL. In addition to the SVM
classifier, we also try the Maximum Entropy (ME) classifier, which
usually gives good accuracy in a wide variety of classification
problems\footnote{We use the logistic regression classifier with
  $L_2$ regularization provided by the \texttt{scikit-learn} software
  package. The regularization term is fixed at 1.}.
Table~\ref{tab:per3} shows the $F_1$ scores of different labelling
strategies.

\begin{table}[!h]
\centering
\caption{Accuracy of two labelling strategies}\label{tab:per3}
\begin{small}
\begin{tabular}{lcc}
\hline\hline
  & ME & SVM \\ 
\hline
 1-step strategy & 69.79\% & 69.96\%  \\ 
 2-step strategy & 69.28\% & 69.38\% \\ 
\hline\hline
 \end{tabular}
\end{small} 
\end{table}

We see that the performance of SVM classifier is slightly better than the performance of ME classifier. The best accuracy is obtained by using 1-step strategy
with SVM classifier. The current SRL system achieves an $F_1$ score of
69.96\%.

\subsubsection{Feature Analysis}
In the fourth experiment, we analyse and evaluate the impact of each
individual feature to the accuracy of our system so as to find the
best feature set for our Vietnamese SRL system. We start with the basic
feature set presented previously, denoted by $\Phi_0$ and augment it
with modified and new features as shown in Table~\ref{tab:fs-1}. The
accuracy of these feature sets are shown in Table~\ref{tab:per4-1}.

\begin{table}[!h]
\centering
\caption{Feature sets}\label{tab:fs-1}
\begin{small}
\begin{tabular}{cl}
\hline\hline
Feature Set & Description \\ 
\hline
$\Phi_{1}$ & $\Phi_{0} \cup $\{Function Tag\} \\ 
$\Phi_{2}$ & $\Phi_{0} \cup $\{Predicate Type\} \\ 
$\Phi_{3}$ & $\Phi_{0 }\cup $\{Distance\} \\ 
\hline\hline
\end{tabular}
\end{small}
\end{table} 

\begin{table}[!h]
\centering
\caption{Accuracy of feature sets in Table~\ref{tab:fs-1}}\label{tab:per4-1}
\begin{small}
\begin{tabular}{cccc}
\hline\hline
Feature Set & Precision & Recall & $F_1$ \\ 
\hline
$\Phi_{0}$ & 73.02\% & 67.16\% & 69.96\% \\ 
$\Phi_{1}$ & \textbf{77.38\%} & \textbf{71.20\%} &  \textbf{74.16\%} \\ 
$\Phi_{2}$ & 72.98\% & 67.15\% & 69.94\% \\ 
$\Phi_{3}$ & \textbf{73.04\%} & \textbf{67.21\%} & \textbf{70.00\%} \\ 
\hline\hline
\end{tabular}
\end{small} 
\end{table}

We notice that amongst the three features, function tag is the most
important feature which increases the accuracy of the baseline feature
set by about 4\% of $F_1$ score. The distance feature also helps
increase slightly the accuracy. We thus consider the fourth feature
set $\Phi_4$ defined as
\begin{equation*}
  \Phi_4 = \Phi_0 \cup \{\text{Function  Tag}\} \cup \{\text{Distance}\}.
\end{equation*}

In the fifth experiment, we investigate the significance of individual
features to the system by removing them, one by one from the feature set
$\Phi_4$. By doing this, we can evaluate the importance of each
feature to our overall system. The feature sets and their corresponding
accuracy are presented in Table~\ref{tab:fs-3} and
Table~\ref{tab:per4-3} respectively.

\begin{table}[!h]
\centering
\caption{Feature sets (continued)} \label{tab:fs-3}
\begin{small}
\begin{tabular}{cl}
\hline\hline
Feature Set & Description \\ 
\hline
$\Phi_{5}$ & $\Phi_{4} \setminus$\{Function Tag\} \\ 
$\Phi_{6}$ & $\Phi_{4}\setminus$\{Distance\} \\ 
$\Phi_{7}$ & $\Phi_{4}\setminus$\{Head Word\} \\ 
$\Phi_{8}$ & $\Phi_{4}\setminus$\{Path\} \\ 
$\Phi_{9}$ & $\Phi_{4}\setminus$\{Position\} \\ 
$\Phi_{10}$ & $\Phi_{4}\setminus$\{Voice\} \\ 
$\Phi_{11}$ & $\Phi_{4}\setminus$\{Subcategorization\} \\ 
$\Phi_{12}$ & $\Phi_{4}\setminus$\{Predicate\} \\
$\Phi_{13}$ & $\Phi_{4}\setminus$\{Phrase Type\} \\
\hline\hline
\end{tabular}
\end{small}
\end{table}

\begin{table}[!h]
\centering
\caption{Accuracy of feature sets in Table~\ref{tab:fs-3}}\label{tab:per4-3}
\begin{small}
\begin{tabular}{cccc}
\hline\hline
Feature Set & Precision & Recall & $F_1$ \\ 
\hline
$\Phi_{4}$ & 77.53\% & 71.29\% &  74.27\% \\ 
$\Phi_{5}$ & 73.04\% & 67.21\% &  70.00\% \\ 
$\Phi_{6}$ & 77.38\% & 71.20\% &  74.16\% \\ 
$\Phi_{7}$ & 73.74\% & 67.17\% &  70.29\% \\ 
$\Phi_{8}$ & 77.58\% & 71.10\% & 74.20\% \\ 
$\Phi_{9}$ & 77.39\% & 71.39\% & 74.26\% \\ 
$\Phi_{10}$ & 77.51\% & 71.24\% & 74.24\% \\ 
$\Phi_{11}$ & \textbf{77.53\%} & \textbf{71.46\%} & \textbf{74.37\%} \\ 
$\Phi_{12}$ & 77.38\% & 71.41\% & 74.27\% \\ 
$\Phi_{13}$ & 77.86\% & 70.99\% & 74.26\% \\ 
\hline\hline
\end{tabular} 
\end{small}
\end{table}

We see that the accuracy increases slightly when the subcategorization feature
($\Phi_{11}$) is removed. For this reason, we remove
only the subcategorization feature. The best feature set includes the
following features: predicate, phrase type, function tag,
parse tree path, distance, voice, position and head word. The
accuracy of our system with this feature set is 74.37\% of $F_1$ score.

\subsubsection{Improvement via Integer Linear Programming}

\begin{table}[!h]
\centering
\caption{The impact of ILP}\label{tab:per4-4}
\begin{small}
\begin{tabular}{lccc}
\hline\hline
 & Precision & Recall & $F_1$ \\ 
\hline
A & 77.53\% & 71.46\% & 74.37\% \\
B & 78.28\% & 71.48\% &  74.72\%\\  
C & 78.29\% & 71.48\% &  \textbf{74.73\%}\\ 
\hline
\multicolumn{4}{l}{A: Without ILP} \\
\multicolumn{4}{l}{B: With ILP (not using constraint 5)} \\
\multicolumn{4}{l}{C: With ILP (using constraint 5)} \\
\hline\hline
\end{tabular}
\end{small}
\end{table}

As discussed previously, after classifying the arguments, we use ILP
method to help improve the overall accuracy. In the sixth experiment,
we set up an ILP to find the best performance satisfying constraints
presented earlier\footnote{We use the GLPK solver provided by the
  \texttt{PuLP} software package, available at
  \url{https://pythonhosted.org/PuLP/}.}. The score
$p_{ic}=score(S^{i}=c)$ is the signed distance of that argument to
the hyperplane. We also compare our ILP system with
  the ILP method for English by using only constraints from 1 to 4.
The improvement given by ILP is shown in 
Table~\ref{tab:per4-4}. We see that ILP increases the performance of
about 0.4\% and when adding constraint 5, the result is slightly
better. The accuracy of for each argument is shown in
Table~\ref{tab:per4-8}.

\begin{table}[!h]
\centering
\caption{Accuracy of each argument type}\label{tab:per4-8} 
\begin{small}
\begin{tabular}{lccc}
\hline\hline
 & Precision & Recall & $F_1$ \\ 
\hline
Arg0 & 93.92\% & 97.34\% & 95.59\% \\ 
Arg1 & 68.97\% & 82.38\% & 75.03\% \\ 
Arg2 & 56.87\% & 46.62\% & 50.78\% \\ 
Arg3 & 3.33\% & 5.00\% & 4.00\% \\ 
Arg4 & 61.62\% & 22.01\% & 31.17\% \\ 
ArgM-ADJ & 0.00\% & 0.00\% & 0.00\% \\ 
ArgM-ADV & 60.18\% & 44.80\% & 51.17\% \\ 
ArgM-CAU & 61.96\% & 47.63\% & 50.25\% \\ 
ArgM-COM & 41.90\% & 78.72\% & 52.53\% \\ 
ArgM-DIR & 41.21\% & 23.01\% & 29.30\% \\ 
ArgM-DIS & 60.79\% & 56.37\% & 58.25\% \\ 
ArgM-DSP & 0.00\% & 0.00\% & 0.00\% \\ 
ArgM-EXT & 70.10\% & 77.78\% & 73.19\% \\ 
ArgM-GOL & 0.00\% & 0.00\% & 0.00\% \\ 
ArgM-I & 0.00\% & 0.00\% & 0.00\% \\ 
ArgM-LOC & 59.26\% & 75.56\% & 66.21\% \\ 
ArgM-LVB & 0.00\% & 0.00\% & 0.00\% \\ 
ArgM-MNR & 56.06\% & 52.00\% & 53.70\% \\ 
ArgM-MOD & 76.57\% & 84.77\% & 80.33\% \\ 
ArgM-NEG & 85.21\% & 94.24\% & 89.46\% \\ 
ArgM-PRD & 22.00\% & 13.67\% & 15.91\% \\ 
ArgM-PRP & 70.38\% & 70.96\% & 70.26\% \\ 
ArgM-Partice & 38.76\% & 17.51\% & 22.96\% \\ 
ArgM-REC & 45.00\% & 48.00\% & 45.56\% \\ 
ArgM-RES & 2.00\% & 6.67\% & 9.52\% \\ 
ArgM-TMP & 78.86\% & 93.09\% & 85.36\% \\ 
\hline\hline
\end{tabular}
\end{small}
\end{table}

A detailed investigation of our constituent extraction algorithm reveals
that it can account for about 86\% of possible argument
candidates. Although this coverage ratio is relatively high, it is not
exhaustive. One natural question to ask is whether an exhaustive
search of argument candidates could improve the accuracy of the system
or not. Thus, in the seventh experiment, we replace our constituent
extraction algorithm by an exhaustive search where all nodes of a
syntactic tree are taken as possible argument candidates. Then, we add
the third constraint to the ILP post-processing step as presented
above (\textit{Arguments cannot overlap other arguments in the
  sentence}). An accuracy comparison of two constituent extraction algorithms 
is shown in Table~\ref{tab:per4-5}. 

\begin{table}[!h]
\centering
\caption{Accuracy of two extraction algorithms}\label{tab:per4-5}
\begin{small}
\begin{tabular}{lcc}
\hline\hline
 & Getting All & Our Extraction \\ 
 & Nodes & Alg. \\ 
\hline
Precision & 19.56\% & 82.15\% \\ 
Recall & \textbf{93.25}\% & 86.12\% \\ 
$F_1$ & 32.23\% & 84.08\% \\ 
\hline\hline
\end{tabular} 
\end{small}
\end{table}
Taking all nodes of a syntactic tree help increase the number of
candidate argument to a coverage ratio of 93.25\%. However, it also
proposes many wrong candidates as shown by a low precision
ratio. Table~\ref{tab:per4-6} shows the accuracy of our system in the
two candidate extraction approaches. 

\begin{table}[!h]
\centering
\caption{Accuracy of our system}\label{tab:per4-6}
\begin{small}
\begin{tabular}{lcc}
\hline\hline
 & Getting All & Our Extraction \\ 
 & Nodes & Alg. \\ 
\hline
Precision & 77.99\% & 78.29\% \\ 
Recall & 62.50\% & 71.48\% \\ 
$F_1$ & 69.39\% & 74.73\% \\ 
\hline\hline
\end{tabular}
\end{small}
\end{table}

We see that an exhaustive search of candidates help present more
possible constituent candidates but it makes the performance of the
system worse than the constituent extraction algorithm (69.39\%
compared to 74.73\% of $F_1$ ratio). One plausible explanation is that
the more a classifier has candidates to consider, the more it is
likely to make wrong classification decision, which results in worse
accuracy of the overall system. In addition, a large number of
candidates makes the system lower to run. In our experiment, we see
the training time increased fourfold when the exhaustive search
approach was used instead of our constituent extraction algorithm.

\subsubsection{Learning Curve}

\begin{figure}[!h]
\begin{center}
\caption{Learning curve of the system}\label{fig:lc}
\begin{small}
\begin{tikzpicture}[y=.06cm, x=.0013cm,font=\sffamily]
	\draw (0,0) -- coordinate (x axis mid) (5000,0);
    	\draw (0,0) -- coordinate (y axis mid) (0,82);
    	\foreach \x in {0,800,...,5000}
     		\draw[yshift=0cm] (\x,1pt) -- (\x,-3pt)
			node[anchor=north] {\x};
    	\foreach \y in {0,10,...,80}
     		\draw (1pt,\y) -- (-3pt,\y) 
     			node[anchor=east] {\y}; 
	\node[below=0.8cm] at (x axis mid) {Number of sentences in training data};
	\node[rotate=0, above=2.6cm] at (y axis mid) {$F_1$};
	
	\draw plot[ mark=*, mark options={fill=black} ]
		file {data.data};
\end{tikzpicture}
\end{small}
\end{center}
\end{figure}
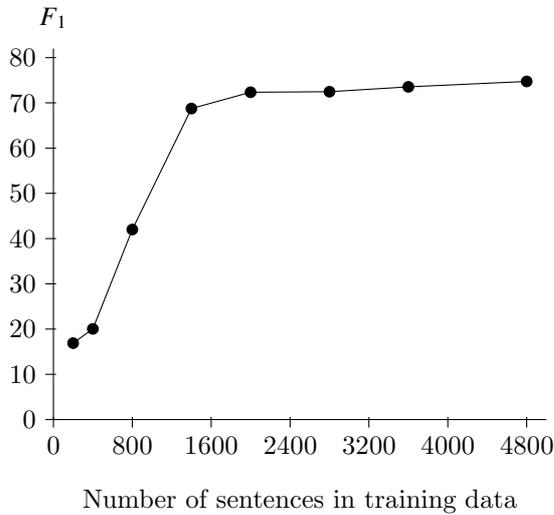

In the ninth experiment, we investigate the dependence of accuracy to
the size of the training dataset. Figure~\ref{fig:lc} depicts the
learning curve of our system when the data size is varied.

It seems that the accuracy of our system improves only slightly
starting from the dataset of about 2,000 sentences. Nevertheless, the
curve has not converged, indicating that the system could achieve a
better accuracy when a larger dataset is available.

\subsection{Generalizing to Unseen Words}
\label{sec:generalizing}

In this section, we report our effort to extend the applicability of
our SRL system to new text domain where rare or
unknown words are common. As seen in the previous systems, some
important features of our SRL system are word features including
predicates and head words. 

As in most NLP tasks, the words are usually encoded as symbolic
identifiers which are drawn from a vocabulary. Therefore, they are
often represented by one-hot vectors (also called indicator vectors)
of the same length as the size of the vocabulary. This representation
suffers from two major problems. The first problem is data sparseness,
that is, the parameters corresponding to rare or unknown words are
poorly estimated. The second problem is that it is not able to capture
the semantic similarity between closely related words. This limitation
of the one-hot word representation has motivated unsupervised methods
for inducing word representations over large, unlabelled corpora.

Recently, distributed representations of words have been shown to be
advantageous for many natural language processing tasks. A distributed
representation is dense, low dimensional and real-valued. Distributed
word representations are called word embeddings. Each dimension of the
embedding represents a latent feature of the word which hopefully captures
useful syntactic and semantic similarities \cite{Turian:2010}.

Word embeddings are typically induced using neural language models,
which use neural networks as the underlying predictive
model. Historically, training and testing of neural language models
has been slow, scaling as the size of the vocabulary for each model
computation \cite{Bengio:2003}. However, many approaches have been
recently proposed to speed up the training process, allowing scaling
to very large
corpora \cite{Morin:2005,Collobert:2008,Mnih:2009,Mikolov:2013a}.

Another method to produce word embeddings has been introduced recently
by the natural language processing group at the Stanford
university \cite{Pennington:2014}. They proposed a global
log-bilinear regression model that combines the advantages of the two
major model families in the literature: global matrix factorization
and local context window methods.

We present in the subsections~\ref{sec:skip} and~\ref{sec:glove} how
we use a neural language model and a global log-bilinear regression
model, respectively, to produce word embeddings for Vietnamese which
are used in this study.

\subsubsection{Skip-gram Model}\label{sec:skip}

We use word embeddings produced by Mikolov's continuous Skip-gram
model using the neural network and source code introduced
in \cite{Mikolov:2013b}. The continuous skip-gram model itself is
described in details in \cite{Mikolov:2013a}.

For our experiments we used a continuous skip-gram window of size 2,
\textit{i.e.} the actual context size for each training sample is a random
number up to 2. The neural network uses the central word in the
context to predict the other words, by maximizing the average
conditional log probability
\begin{equation}
\frac{1}{T} \sum\limits_{t=1}^T \sum\limits_{j=-c}^c \log
p(w_{t+j}|w_t),
\end{equation}
where $\{w_i: i \in T\}$ is the whole training set, $w_t$ is the central
word and the $w_{t+j}$ are on either side of the context. The
conditional probabilities are defined by the softmax function  
\begin{equation}
p(a|b) = \frac{\exp (o_a^\top i_b)}{\sum\limits_{w \in \mathcal V}
  \exp(o_w^\top i_b)},
\end{equation}
where $i_w$ and $o_w$ are the input and output vector of $w$
respectively, and $\mathcal V$ is the vocabulary. For computational efficiency,
Mikolov's training code approximates the softmax function by the
hierarchical softmax, as defined in \cite{Morin:2005}. Here the
hierarchical softmax is built on a binary Huffman tree with one word
at each leaf node. The conditional probabilities are calculated
according to the decomposition:
\begin{equation}
p(a|b) = \prod\limits_{i=1}^l p(d_i(a)|d_1(a)... d_{i-1}(a), b),
\end{equation}
where $l$ is the path length from the root to the node $a$, and
$d_i(a)$ is the decision at step $i$ on the path (for example $0$ if
the next node the left child of the current node, and $1$ if it is the
right child). If the tree is balanced, the hierarchical softmax only
needs to compute around $\log_2 |\mathcal V|$ nodes in the tree, while the
true softmax requires computing over all $|\mathcal V|$ words.

\begin{figure}[!h]
  \centering
  \caption{Some Vietnamese words produced by the Skip-gram model,
    projected onto two dimensions.}
  \label{fig:skip}
  \includegraphics[scale=0.5]{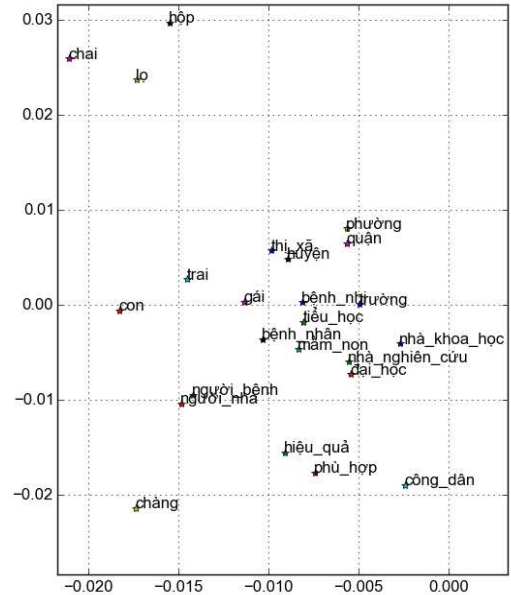}
\end{figure}

The training code was obtained from the tool
\texttt{word2vec}\footnote{\url{http://code.google.com/p/word2vec/}}
and we used frequent word subsampling as well as a word appearance
threshold of 5. The output dimension is set to 50, \textit{i.e.} each
word is mapped to a unit vector in $\mathbb{R}^{50}$. This is deemed
adequate for our purpose without overfitting the training
data. Figure~\ref{fig:skip} shows the scatter plot of some
Vietnamese words which are projected onto the first two principal
components after performing the principal component analysis of all
the word distributed representations. We can see that semantically
related words are grouped closely together.

\subsubsection{GloVe Model}\label{sec:glove}
Pennington, Socher, and Manning \citep{Pennington:2014} introduced the
global vector model for learning word representations (GloVe). Similar to the
Skip-gram model, GloVe is a local context window method but it
has the advantages of the global matrix factorization method. 

The main idea of GloVe is to use word-word occurrence counts to
estimate the co-occurrence probabilities rather than the probabilities by
themselves. Let $P_{ij}$ denote the probability that word $j$ appear
in the context of word $i$; $w_i \in \mathbb R^d$ and $w_j \in \mathbb
R^d$ denote the word vectors of word $i$ and word $j$
respectively. It is shown that 
\begin{equation}
w_i^{\top} w_j = \log(P_{ij}) = \log(C_{ij}) - \log(C_i),
\end{equation}
where $C_{ij}$ is the number of times word $j$ occurs in the context
of word $i$. 

It turns out that GloVe is a global log-bilinear regression
model. Finding word vectors is equivalent to solving a weighted
least-squares regression model with the cost function:
\begin{equation}
  J = \sum_{i,j = 1}^n f(C_{ij})(w_i^{\top} w_j + b_i + b_j - \log(C_{ij}))^2,
\end{equation}
where $n$ is the size of the vocabulary, $b_i$ and $b_j$ are
additional bias terms and $f(C_{ij})$ is a weighting function. A class
of weighting functions which are found to work well can be
parameterized as 
\begin{equation}
f(x) =\begin{cases}
\left(\frac{x}{x_{\max}}\right)^\alpha & \text{if } x < x_{\max} \\
1 & \text{otherwise}
\end{cases}
\end{equation}

\begin{figure}[!h]
  \centering
  \caption{Some Vietnamese words produced by the GloVe model,
    projected onto two dimensions.}
  \label{fig:glove}
  \includegraphics[scale=0.5]{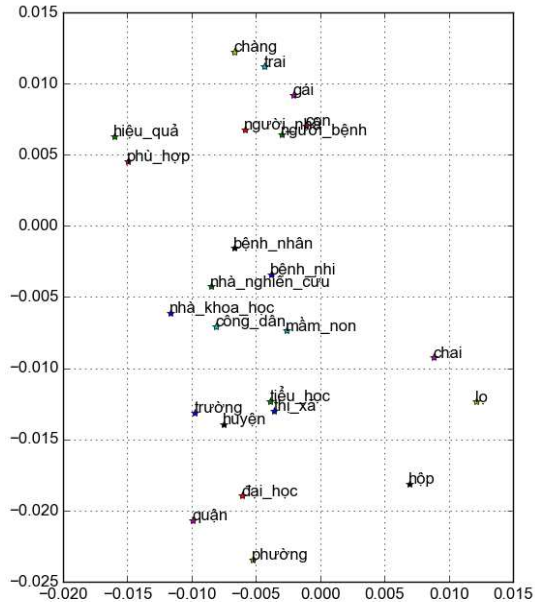}
\end{figure}

The training code was obtained from the tool
\texttt{GloVe}\footnote{\url{http://nlp.stanford.edu/projects/glove/}}
and we used a word appearance threshold of
2,000. Figure~\ref{fig:glove} shows the scatter plot of the same words
in Figure~\ref{fig:skip}, but this time their word vectors are produced
by the GloVe model. 

\subsubsection{Text Corpus} 
To create distributed word representations, we use a dataset
consisting of 7.3GB of text from 2 million articles collected
through a Vietnamese news portal\footnote{\url{http://www.baomoi.com}}. The
text is first normalized to lower case and all special characters are
removed except these common symbols: the comma, the semicolon, the
colon, the full stop and the percentage sign. All numeral sequences
are replaced with the special token \textless number\textgreater,\ so
that correlations between certain words and numbers are correctly
recognized by the neural network or the log-bilinear regression model.

Each word in the Vietnamese language may consist of more than one
syllables with spaces in between, which could be regarded as multiple
words by the unsupervised models. Hence it is necessary to replace the
spaces within each word with underscores to create full word
tokens. The tokenization process follows the method described
in \cite{Le:2008a}.

After removal of special characters and tokenization, the articles add
up to $969$ million word tokens, spanning a vocabulary of $1.5$ million unique
tokens. We train the unsupervised models with the full vocabulary to obtain
the representation vectors, and then prune the collection of word
vectors to the $65,000$ most frequent words, excluding special symbols and
the token \textless number\textgreater \, representing numeral sequences.

\subsubsection{SRL with Distributed Word Representations}

We train the two word embedding models on the same text
corpus presented in the previous subsections to produce distributed
word representations, where each word is represented by a real-valued
vector of 50 dimensions. 

In the last experiment, we replace predicate or head word features in
our SRL system by their corresponding word vectors. For predicates
which are composed of multiple words, we first tokenize them into
individual words and then average their vectors to get vector
representations. Table~\ref{tab:per4-9} and Table~\ref{tab:per4-10}
shows performances of the Skip-gram and GloVe models for predicate
feature and for head word feature, respectively.
\begin{table}[!h]
\centering
\caption{The impact of word embeddings of predicate}\label{tab:per4-9}
\begin{small}
\begin{tabular}{lccc}
\hline\hline
 & Precision & Recall & $F_1$ \\ 
\hline
A & 78.29\% & 71.48\% & 74.73\% \\
B  & 78.37\% & 71.49\% & 74.77\% \\ 
C & 78.29\% & 71.38\% &  74.67\%\\ 
\hline
\multicolumn{4}{l}{A: Predicate word} \\
\multicolumn{4}{l}{B: Skip-gram vector} \\
\multicolumn{4}{l}{C: GloVe vector} \\
\hline\hline
\end{tabular}
\end{small}
\end{table}

\begin{table}[!h]
\centering
\caption{The impact of word embeddings of head word}\label{tab:per4-10}
\begin{small}
\begin{tabular}{lccc}
\hline\hline
 & Precision & Recall & $F_1$ \\ 
\hline
A		& 78.29\% & 71.48\% & 74.73\% \\
B		& 77.53\% & 70.76\% & 73.99\% \\ 
C		& 78.12\% & 71.58\% &  74.71\%\\ 
\hline
\multicolumn{4}{l}{A: Head word} \\
\multicolumn{4}{l}{B: Skip-gram vector} \\
\multicolumn{4}{l}{C: GloVe vector} \\
\hline\hline
\end{tabular}
\end{small}
\end{table}

We see that both of the two types of word embeddings do not decrease
the accuracy of the system. In other words, their use can help
generalize the system to unseen words.

\section{Conclusion}
\label{sec:conclusion}

We have presented our work on developing a semantic role labelling
system for the Vietnamese language. The system comprises two main
component, a corpus and a software. Our system achieves a good
accuracy of about 74.8\% of $F_1$ score. 

We have argued that one cannot assume a good applicability of existing
methods and tools developed for English and other occidental languages
and that they may not offer a cross-language validity. For an
isolating language such as Vietnamese, techniques developed for
inflectional languages cannot be applied ``as is''. In particular, we
have developed an algorithm for extracting argument candidates which
has a better accuracy than the 1-1 node mapping algorithm. We have
proposed some novel features which are proved to be useful for
Vietnamese semantic role labelling, notably and function tags and
distributed word representations. We have employed integer linear
programming, a recent inference technique capable of incorporate a
wide variety of linguistic constraints to improve the performance of
the system. We have also demonstrated the efficacy of distributed word
representations produced by two unsupervised learning models in
dealing with unknown words.

In the future, we plan to improve further our system, in the one hand,
by enlarging our corpus so as to provide more data for the system. On
the other hand, we would like to investigate different models used in
SRL, for example joint models \cite{FitzGerald:2015},
where arguments and semantic roles are jointly embedded in a shared
vector space for a given predicate. In addition, we would like to
explore the possibility of integrating dynamic constraints in the
integer linear programming procedure. We expect the overall
performance of our SRL system to improve.

Our system, including software and corpus, is available as an open source
project for free research purpose and we believe that it is a good
baseline for the development and comparison of future Vietnamese SRL
systems\footnote{\url{https://github.com/pth1993/vnSRL}}. We plan to
integrate this tool to Vitk, an open-source toolkit for processing
Vietnamese text, which contains fundamental processing tools and are
readily scalable for processing very large text
data\footnote{\url{https://github.com/phuonglh/vn.vitk}}.

\section*{Acknowledgement}

We would like to thank Vietnamese linguists at Vietnam Centre of Lexicography
for their collaboration in developing the Vietnamese PropBank. We would also
like to thank the FPT Technology Research Institute for its partial
financial aid. The first author is partly funded by the
Vietnam National University, Hanoi (VNU) under project number
QG.15.04. We are grateful to our anonymous reviewers for their
helpful comments.

\end{document}